\pdfoutput=1

\documentclass[11pt]{article}

\usepackage{acl}

\usepackage{times}
\usepackage{latexsym}
\usepackage{amssymb}
\usepackage{graphicx}
\usepackage{algpseudocode}
\usepackage{algorithm}
\usepackage{xparse}
\usepackage{tabularx,tabulary}
\usepackage{multirow}
\usepackage{nicefrac}
\usepackage{color, colortbl}
\usepackage{adjustbox}
\usepackage{tabularx}
\usepackage{booktabs}
\usepackage{subcaption}
\usepackage{makecell}
\usepackage{arydshln}
\usepackage{tikz}
\usepackage{xcolor}

\usepackage{enumitem}
\makeatletter
\NewDocumentCommand{\LeftComment}{s m}{%
  \Statex \IfBooleanF{#1}{\hspace*{\ALG@thistlm}}\(\triangleright\) #2}
\makeatother

\usepackage[T1]{fontenc}

\usepackage[utf8]{inputenc}

\usepackage{microtype}

\title{\textsc{lad}: Language Models as Data for Zero-Shot Dialog}

\author{Shikib Mehri$^\spadesuit$ \hspace{.75em} Yasemin Altun$^\diamondsuit$ \hspace{.75em} Maxine Eskenazi$^\spadesuit$ \\ 
  $^\spadesuit$ Carnegie Mellon University \hspace{.75em}  $^\diamondsuit$ Google \\
  \texttt{amehri@cs.cmu.edu,altun@google.com,max@cs.cmu.edu} \\}

\begin{document}
\maketitle
\begin{abstract}
To facilitate zero-shot generalization in task-oriented dialog, this paper proposes \textit{Language Models as Data} (\textsc{lad}). \textsc{lad} is a paradigm for creating \textit{diverse} and \textit{accurate} synthetic data which conveys the necessary structural constraints and can be used to train a downstream neural dialog model. \textsc{lad} leverages GPT-3 to induce linguistic diversity. \textsc{lad} achieves significant performance gains in zero-shot settings on intent prediction (\textbf{+15\%}), slot filling (\textbf{+31.4 F-1}) and next action prediction (\textbf{+11 F-1}). Furthermore, an interactive human evaluation shows that training with \textsc{lad} is competitive with training on human dialogs. \textsc{lad} is open-sourced, with the code and data available at \textbf{\url{https://github.com/Shikib/lad}}.
\end{abstract}

\section{Introduction}

A long-standing goal of dialog research is to develop mechanisms for flexibly adapting dialog systems to new domains and tasks \citep{rastogi2020towards,mosig2020star}. While the advent of large-scale pre-training \citep{devlin2018bert,liu2019roberta,zhang2019dialogpt} has brought about significant progress in few-shot and zero-shot generalization across many different problems in Natural Language Processing \citep{brown2020language,wei2021finetuned}, zero-shot generalization in \textbf{task-oriented dialog} remains elusive. A likely reason for this discrepancy is that dialog models require significant data because they need to learn task-specific \textbf{structural constraints}, such as the domain ontology and the dialog policy. While large language models (e.g., GPT-3) exhibit strong language understanding and generation abilities \citep{brown2020language}, they have no \textit{a priori} knowledge of the structural constraints implied by a specific (unseen) problem setting (e.g., relevant intents, dialog policy, etc.). As such, in order to adapt a pre-trained LM for task-oriented dialog, it is necessary to \textit{impose \textbf{structural constraints} on the \textbf{unstructured} representation space of a pre-trained model}. Fine-tuning moderately-sized language models (LMs) (e.g., BERT) with well-motivated inductive biases \citep{mitchell1980need} facilitates sample-efficient learning of the structural constraints \citep{peng2020soloist,henderson2020convex,mehri2021schema}. However, fine-tuning can be impractical (e.g., in academic settings) with large LMs (e.g., GPT-3) due to the cost, computational power and immutable architectures. To this end, this paper aims to address the following: \textit{`How can we leverage the strong language understanding and generation abilities of large LMs to facilitate \textbf{zero-shot generalization} in task-oriented dialog?'}  

\begin{figure}
    \centering
    \includegraphics[width=0.5\textwidth]{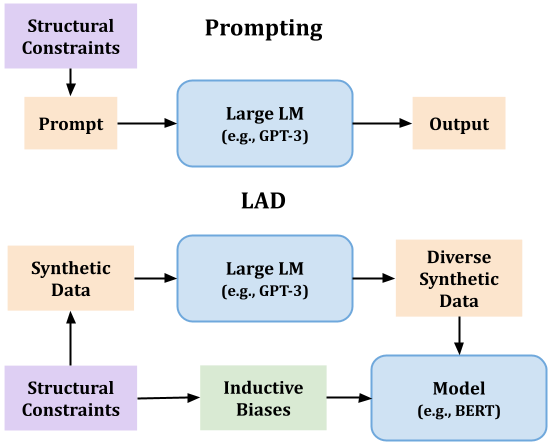}
    \caption{Prompting must convey the structural constraints through a natural language prompt. In contrast, \textsc{lad} uses large LMs to induce diversity in a synthetic dataset. As such, \textsc{lad} conveys structural constraints through both the synthetic data and the inductive biases in the downstream problem-specific models.}
    \label{fig:my_label}
\end{figure}

Given the in-context meta-learning abilities of large LMs \citep{brown2020language}, prior work has explored prompt-engineering or prompt-tuning \citep{reynolds2021prompt,lester2021power,madotto2021few}. Well-designed prompts can convey the necessary structural constraints. However, it is challenging to express complex constraints (e.g., a dialog policy) in natural language. Prompting also precludes inductive biases in the model (architecture, training algorithm, etc.) and over-relies on the meta-learning abilities of large LMs. As such, there is a tradeoff between prompting large LMs (i.e., generalizable NLU and NLG) and fine-tuning smaller LMs (i.e., problem-specific inductive biases, efficiency). A potential interpretation for the strength of large LMs is that they learn the distributional structure of language \citep{harris1954distributional} by observing web-scale data \citep{sinha2021masked}. Motivated by this interpretation, this paper proposes \textit{Language Models as Data} (\textbf{\textsc{lad}}). 

\textsc{\textbf{lad}} is a novel paradigm in which large LMs are used in a zero-shot domain-agnostic manner to induce \textit{linguistic diversity} in synthetic data. Given a \textit{minimal expression}\footnote{A minimal expression can be defined as the \textit{smallest} amount of data necessary to express a structural constraint. For example, one utterance to define an intent class.} of the structural constraints (henceforth referred to as a \textbf{schema}), \textsc{lad} (1) creates a seed synthetic dataset using domain-agnostic algorithms, (2) leverages large LMs to \textit{reformulate} utterances, and (3) validates the resulting data to ensure adherence to the schema. The resulting synthetic data, which is sufficiently \textbf{diverse} and expresses the necessary \textbf{structural constraints}, can be used to train neural dialog models. In contrast to prompting, \textsc{lad} facilitates zero-shot generalization by (1) leveraging the sophisticated abilities of large LMs (knowledge of the distributional structure of language) to induce \textit{linguistic diversity} in the synthetic data while (2) maintaining inductive biases (motivated by the structural constraints) in the problem-specific model architectures.

The challenge of creating synthetic data that is indistinguishable from human-annotated data, both in its expression of structural constraints and in its diversity, is highly impractical \citep{lin2021domain,feng2021survey}. Instead, the goal of this work is to create synthetic data that is \textit{sufficient} to train a sample-efficient and robust model. Therefore, the claim of this paper is that \textsc{lad} can create synthetic data, conditioned on a minimal expression of structural constraints (i.e., a schema), that can be used to train robust and sample-efficient neural models and induce performance gains in zero-shot settings.




To validate this claim, \textsc{lad} is applied to three problems in dialog: intent prediction, slot filling and next action prediction. Next action prediction is particularly difficult in zero-shot settings since the structural constraints include the \textit{dialog policy}. \textsc{lad} demonstrates significant gains across five datasets (\textbf{+10} to \textbf{+30} improvements on F-1 and accuracy) in zero-shot settings when evaluating on human-annotated corpora. To further validate the efficacy of \textsc{lad}, an interactive evaluation with humans (over 1600 dialogs) is performed. The results of this interactive evaluation suggest that \textsc{lad} can yield performance comparable to training on human dialogs. The claim of this paper is validated empirically across multiple datasets. \textsc{lad} is shown to generate diverse and accurate synthetic data, which is subsequently used to train neural dialog models and facilitate zero-shot generalization.

\section{Definitions}

Zero-shot generalization can be conceptualized as \textit{imposing \textbf{structural constraints} on the unstructured representation space of a pre-trained model, using a given \textbf{schema} (i.e., minimal expression).} We begin with a neural network, $\mathcal{M}$, with general language understanding abilities and limited knowledge of task-oriented dialog (e.g., BERT \citep{devlin2018bert}). The necessary \textbf{structural constraints} that $\mathcal{M}$ must learn are implied by the target dialog setting, i.e., the problem (e.g., next action prediction), the domain (e.g., restaurants) and the task (e.g., restaurant reservation). These structural constraints conceptually define the desired properties for the representations of $\mathcal{M}$, i.e., \textit{what must be learned} by $\mathcal{M}$. In a full-shot setting, the constraints are conveyed by a human-annotated dataset and thereby learned through supervised learning. In contrast, the goal in zero-shot generalization is to learn these structural constraints from a minimal expression, i.e., a \textbf{schema}. The following sections formally define structural constraints and schemas.

Throughout this paper, \textbf{zero-shot} refers to a setting wherein the only \textit{human-annotated} data is the schema. Since a schema is a minimal expression of the necessary structural constraints, we argue that it is impossible to use less data, without making assumptions about the prior knowledge of a pre-trained model. Such assumptions would limit the generality of a method for zero-shot generalization.





\subsection{Structural Constraints}


To effectively adapt a model, particularly in zero-shot settings, it is imperative to define \textit{what the model must learn}. Structural constraints conceptualize the desired properties for the representations of a model $\mathcal{M}$. Understanding these structural constraints allows us to design an effective paradigm to facilitate zero-shot generalization. Concretely, knowledge of the structural constraints influences (1) the inductive biases \citep{mitchell1980need} in the model architecture, (2) the design of the schema, and (3) the algorithms used to create synthetic data.


\textbf{Intent prediction}, for example, is the problem of classifying an utterance $u \in \mathcal{U}$ to an intent $i \in \mathcal{I}$. An intent prediction model $\mathcal{M}_{I}$ must learn to produce similar representations $\mathcal{M}_{I}(u)$ for all utterances that have the same intent. Learning this structural constraint is equivalent to transforming the unstructured representation space of $\mathcal{M}$ to the structured output space (i.e., the intent classes).



In the problem of \textbf{slot filling}, for a given utterance $u = \{w_1,w_2,\dots, w_n \}$ and a slot key $s \in \mathcal{S}$, we must predict the corresponding slot value for $s$. The value will either be a contiguous span from $u$, $w_{i:i+k}$, or $\mathrm{none}$. A slot filling model $\mathcal{M}_S$ must learn two sets of structural constraints. First, the representation of $u$ (or the contextual representation of $w \in u$) must follow the structural constraints of intent prediction. Second, each slot value representation $\mathcal{M}_S(w_{i:i+k})$ should be similar to other values for the slot $s$. These two constraints impose structure on both the utterance-level and the span-level representations of $\mathcal{M}_S$.


The structural constraints of intent prediction and slot filling are straightforward and are often learned by a linear layer in supervised settings \citep{casanueva2020efficient,mehri2020dialoglue}. The constraints for the problem of \textbf{next action prediction} are more complex. Next action prediction is the problem of predicting the next system action $a \in \mathcal{A}$ conditioned on the dialog history $u_1, u_2, \dots, u_n$ according to some dialog policy. Given the intents and slots in the dialog history, $\mathcal{I}_D = \{ i_1, i_2, \dots, i_m\}$ and $\mathcal{S}_D = \{ s_1, s_2, \dots, s_k\}$, the dialog policy can be expressed as a function of these intents and slots, $a = \mathrm{policy}(\mathcal{I}_D, \mathcal{S}_D)$. As such a next action prediction model $\mathcal{M}_A$ must learn (1) the structural constraints of intent prediction, (2) of slot filling and (3) the mapping defined by the $\mathrm{policy}$ function. The complexity of third constraint led to the schema-guided paradigm \citep{mehri2021schema}, wherein the policy is explicitly expressed rather than being learned implicitly.

\subsection{Schema}

While structural constraints conceptualize what a model $\mathcal{M}$ must learn, the schema is a minimal expression of these constraints. Imagine that our objective is to train a human (i.e., $\mathcal{M}$ with human-level language understanding and reasoning abilities) to perform task-oriented dialog. Structural constraints define \textit{what} the human must learn. The schema is the \textit{minimum} amount of information needed, for the human to learn the necessary structural constraints, without prior knowledge.

For \textbf{intent prediction}, we define the schema to be a single utterance $u$ for each intent $i \in \mathcal{I}$. \textbf{Slot filling} similarly relies on one utterance $u$ for each slot type $s \in \mathcal{S}$. However, this one utterance only conveys the first structural constraint of slot filling. To ensure that $\mathcal{M}_S$ can learn meaningful span-level representations, the schema for slot filling also includes multiple\footnote{While the number of slot value examples could potentially reduced to 1, up to 20 are used in this paper.} examples of values for each slot.

\textbf{Next action prediction} has three constraints. The first two constraints are equivalent to those of intent prediction and slot filling. As such, the schema includes both (1) one utterance for each intent and (2) a set of slot values for each slot type. To express the structural constraints of the dialog policy, we leverage the graph-based representations of the task-specific dialog policy proposed by \citet{mosig2020star} and \citet{mehri2021schema}.

\section{\textsc{lad}: Language Models as Data} 

\begin{figure*}
    \centering
    \includegraphics[width=\textwidth]{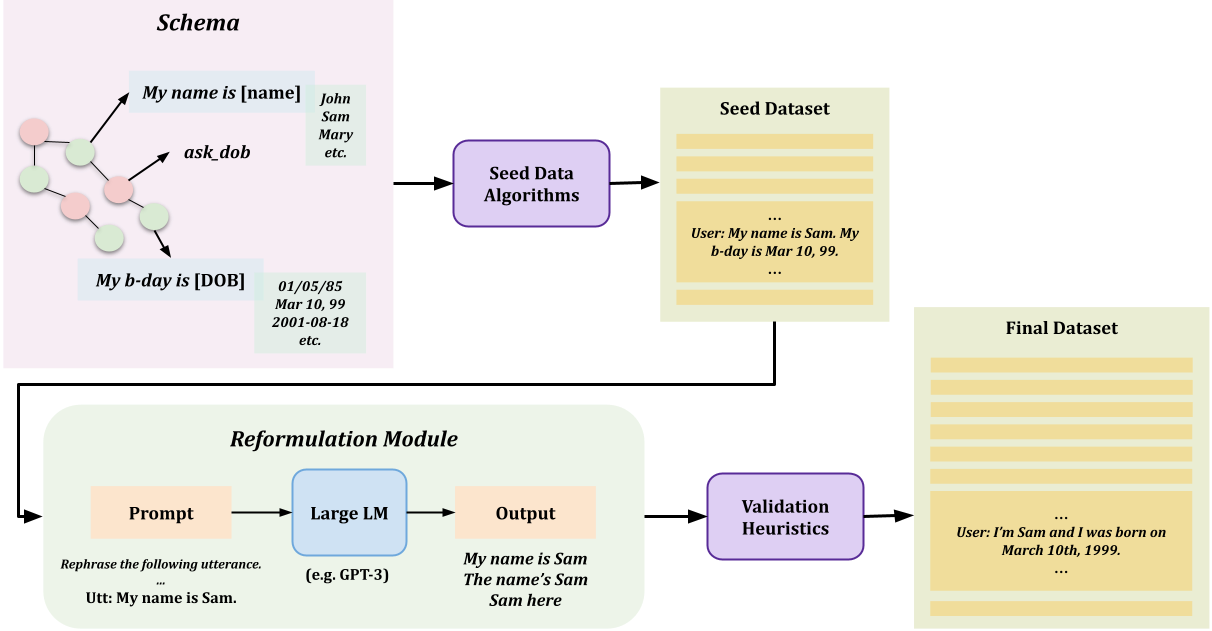}
    \caption{Visualization of \textsc{lad}. (1) Domain-agnostic algorithms use the schema to create a seed dataset which conveys the necessary structural constraints. (2) Large LMs reformulate individual utterances to add linguistic diversity. (3) Validation heuristics are used to ensure adherence to the schema.}
    \label{fig:sgsdc2}
\end{figure*}


Despite exhibiting strong language understanding and generation abilities \citep{brown2020language}, large LMs have no \textit{a priori} knowledge of the structural constraints of task-oriented dialog. Furthermore, imposing the necessary structural constraints on large LMs is impractical due to (1) the difficulty of fine-tuning (cost, computation, immutable architectures) and (2) the limitations of natural language prompts. As such, \textit{Language Models as Data} (\textbf{\textsc{lad}}) uses GPT-3 \citep{brown2020language} to generate \textbf{diverse} synthetic data that express the necessary task-specific \textbf{structural constraints} and can therefore be used to train neural dialog models.

\textbf{\textsc{lad}} is a framework for inducing zero-shot generalization in task-oriented dialog by creating \textit{diverse} and \textit{accurate} synthetic data. \textsc{lad}, visualized in Figure \ref{fig:sgsdc2}, is a three step process: (\S\ref{seed}) domain-agnostic algorithms generate a \textit{seed dataset} from a schema, (\S\ref{reformulate}) GPT-3 \textit{reformulates} utterances in order to induce linguistic diversity, (\S\ref{validate}) heuristics are used to \textit{validate} the reformulated data to ensure adherence to the schema. \textsc{lad} facilitates zero-shot generalization by explicitly leveraging the strengths of large LMs (knowledge of the distributional structure of language) without sacrificing the inductive biases (motivated by structural constraints) in the downstream neural dialog models.

\subsection{Seed Data Creation}
\label{seed}
\textsc{lad} begins by creating seed synthetic data from a given schema. This is a domain-agnostic process that aims to generate synthetic data which accurately convey the necessary structural constraints.

For \textbf{intent prediction}, the schema consists of one utterance for each intent class (sampled from the original corpus) and is used as the seed dataset. For \textbf{slot filling}, the schema consists of one manually-written template utterance and multiple slot values for each slot type. To construct the seed data: (1) begin with the utterance templates from the schema (e.g., \textit{`My first name is \{first\_name\}'}), (2) exhaustively combine template utterances to ensure coverage of slot type combinations, and (3) fill slot values by sampling from the schema. 


The relative complexity of the structural constraints for \textbf{next action prediction}, particularly the dialog policy, necessitates a more sophisticated algorithm for generating the seed data. In order to avoid over-fitting and to ensure that the structural constraints are effectively learned by the model, it is imperative that the synthetic data produced by \textsc{lad} be diverse and realistic. While linguistic diversity is induced through the reformulation with GPT-3, the synthetic dialogs created for next action prediction must also exhibit diversity of \textit{user behavior}. The dialog policy expressed by the schema deterministically defines the \textit{system} behavior. However, users should be able to deviate from the policy, e.g. by providing information out of turn. To account for this, Algorithm \ref{alg:cap} in the Appendix generates a dialog by traversing the dialog policy graph and randomly combining multiple template utterances (e.g., \textit{`\textbf{System:} What is your name? \textbf{User:} My name is John. My phone number is...'}). 



\subsection{Reformulation}
\label{reformulate}
To ensure that downstream neural dialog models can effectively learn the structural constraints, it is imperative that the synthetic data is sufficiently diverse. The seed synthetic data is formulaic and artificial: (1) there is a single template utterance for each user action and (2) when multiple user actions are combined they are simply concatenated. As such, the goal of the reformulation step is two-fold: (1) to induce linguistic diversity and (2) to rephrase concatenations of disjoint template utterances (\textit{`My name is Sarah. I want to plan a party. The day should be Sunday'}) into a natural utterance (\textit{`I'm Sarah and I'd like to plan a party for Sunday.'}).

To reformulate utterances in a domain-agnostic manner, \textsc{lad} leverages the in-context meta-learning abilities of GPT-3 \citep{brown2020language}. Through manual experimentation in the OpenAI Playground\footnote{https://beta.openai.com/playground}, an appropriate prompt is constructed. The prompt begins with an instruction (\textit{`Given a set of sentences, generate 5 natural utterances that convey the same meaning.'}) and includes six examples (details can be found in the Appendix).

Rather than producing a single reformulation of the input, the chosen prompt instructs GPT-3 to generate \textbf{five} utterances. Through the examples provided in the prompt, GPT-3 learns that it should produce five \textit{diverse} reformulations. As such, linguistic diversity is induced through both the decoding algorithm and the six examples in the prompt.






\subsubsection{Scalability}

The cost of the GPT-3 API is approximately \$0.05 USD per reformulation. In order to generate a substantial amount of synthetic data without incurring significant costs, the reformulation step of \textsc{lad} must be performed in a scalable manner. The seed utterances are grouped by their intents and slot keys (e.g., `name;date;time', `name;date', `date;time'). A subset of utterances in each group is reformulated. These reformulated utterances are used as templates and the slot values are randomly replaced. In this manner, the cost scales with respect to the number of distinct intent/slot combinations rather than the desired size of the synthetic dataset. 

\subsection{Validation}
\label{validate}
The seed data will always adhere to the schema and therefore \textit{accurately} convey the necessary structural constraints. However, the reformulated utterances may not be accurate. GPT-3 may modify the intended meaning of an input utterance, for example by ignoring certain slot values. To ensure that the structural constraints are accurately expressed in the final dataset, the reformulation step of \textsc{lad} filters out erroneous reformulations. For slot filling and next action prediction, this is done by ensuring that all of the slot values present in the original utterance (from the seed dataset) are also present in the reformulated utterances (produced by GPT-3).



\subsection{Dataset Statistics}

\textsc{lad} is evaluated on five different datasets. For intent prediction, Banking77 \citep{casanueva2020efficient}, CLINC150 \citep{larson-etal-2019-evaluation}, and HWU64 \citep{liu2019benchmarking} are used. For slot filling, Restaurant8k \citep{coope2020span} is used. For next action prediction, STAR \citep{mosig2020star} is used. Given a human-annotated corpus, a schema is created to express the necessary constraints. \textsc{lad} is then leveraged to create a synthetic dataset conditioned on the schema. Table \ref{tab:dataset_stats} describes the size and creation cost of each of the synthetic datasets.

\begin{table}[]
\renewcommand*{\arraystretch}{1.1}
\small
    \centering
{%
\begin{tabularx}{0.5\textwidth}{@{\extracolsep{\stretch{1}}} l *{3}{c}}
\toprule

       \textbf{Original Dataset} & \textbf{Seed} & \textbf{\textsc{lad}} &  \textbf{Cost (USD)}\\ \midrule
\multicolumn{4}{c}{\textbf{Intent Prediction}} \\ \midrule

HWU64 (8955) & 64 & 800 & \$19 \\
CLINC150 (15000) & 150 & 1664 & \$43 \\
Banking77 (8633) & 77 & 848 & \$25 \\ \midrule

\multicolumn{4}{c}{\textbf{Slot Filling}} \\ \midrule

Restaurant8k (8633) & 85 & 32000 & \$89 \\ \midrule

\multicolumn{4}{c}{\textbf{Next Action Prediction}} \\ \midrule

STAR (1200) & 24000 & 22327 & \$226 \\ \bottomrule
\end{tabularx}
}
    \caption{Statistics for the synthetic datasets created by \textsc{lad}. This table lists the size of the original dataset, the seed dataset and the final synthetic dataset produced by \textsc{lad}. The last column indicates the approximate cost of using GPT-3 for each of the datasets.}
    \label{tab:dataset_stats}
\end{table}




\section{Experiments} 

This paper claims that \textsc{lad} can use a schema to create a sufficiently diverse and accurate synthetic dataset, which can be used to train neural dialog models and facilitate performance gains in zero-shot settings. To validate this claim, experiments are carried out on intent prediction, slot filling and next action prediction across five datasets. 

For each problem, an appropriate model from prior work is identified. The chosen models (1) exhibit strong zero-shot and few-shot generalizability, and (2) are open-source. Though \textsc{lad} is not guaranteed to produce \textit{perfectly} accurate and diverse data, the inductive biases in the chosen models make them more robust to potential errors and limitations in the synthetic data.

\begin{table*}[!t]
\centering
\renewcommand*{\arraystretch}{1.1}
{
\begin{tabularx}{\linewidth}{l X  X  X}
\toprule
  \textbf{Model (Training Data)} & \multicolumn{1}{c}{\bf \textsc{banking77}} & \multicolumn{1}{c}{\bf \textsc{clinc150}} & \multicolumn{1}{c}{\bf \textsc{hwu64}} \\
  \cmidrule(lr){1-1} \cmidrule(lr){2-2} \cmidrule(lr){3-3} \cmidrule(lr){4-4}

 &    & &  \\[-2ex] 
 
\textsc{CBEO (one-shot)}  & \multicolumn{1}{c}{31.36}   & \multicolumn{1}{c}{53.96}  & \multicolumn{1}{c}{43.12} \\

\textsc{CBEO (one-shot + \textsc{lad})}  & \multicolumn{1}{c}{\textbf{51.17}}   & \multicolumn{1}{c}{\textbf{68.11}}  & \multicolumn{1}{c}{\textbf{65.50}} \\

\cmidrule(lr){1-4}

\textsc{CBEO (full-shot)} & \multicolumn{1}{c}{93.83} &  \multicolumn{1}{c}{97.31} & \multicolumn{1}{c}{93.03} \\
\bottomrule
\end{tabularx}}%
\caption{Experimental results on intent prediction. We report the accuracy of training \textsc{CBEO} on (1) one utterance/intent (i.e., the seed data) and (2) the synthetic data produced by \textsc{lad}. For reference, we also show the results reported by \citet{mehri2021example} obtained with full human-annotated training datasets. }
\label{tab:intent}
\end{table*}

\subsection{Intent Prediction}

\textsc{ConvBERT}+\textit{Example-Driven+Observers} (\textsc{CBEO}) \citep{mehri2021example} is used for intent prediction. \textsc{CBEO} learns to predict utterance intents by explicitly comparing to a set of examples. Predicting intents through an explicit non-parametric comparison to examples is an inductive bias that facilitates sample-efficient learning of the structural constraints.

The experimental results shown in Table \ref{tab:intent} demonstrate that the synthetic data produced by \textsc{lad} significantly increase performance on one-shot\footnote{This setting is characterized as one-shot since the utterances in the schema are sampled from the respective dataset.} intent prediction. \textsc{lad} facilitates \textbf{15\%+} accuracy improvement across all three intent prediction datasets. For intent prediction, \textsc{lad} does not use any heuristics during the creation of the seed data or during the validation step. As such, these improvements can be attributed to the reformulation step, which leverages the prompt-driven generation abilities of GPT-3 \citep{brown2020language}. 



\subsection{Slot Filling}

\begin{table}[!t]
\centering
\renewcommand*{\arraystretch}{1.1}
{
\begin{tabularx}{\linewidth}{l X }
\toprule
  \textbf{Model} & \multicolumn{1}{c}{\bf \textsc{F-1}} \\
  
  &  \\[-2ex] 
 
  \cmidrule(lr){1-2}  \multicolumn{2}{c}{\textbf{Zero-Shot Results}} \\   \cmidrule(lr){1-2}

\textsc{ConVEx {\small \citep{henderson2020convex}}}  &  \multicolumn{1}{c}{5.2} \\

\textsc{Coach+TR {\small \citep{liu2020coach}}}  &  \multicolumn{1}{c}{10.7} \\

\textsc{GenSF {\small \citep{mehri2021gensf}}}  &  \multicolumn{1}{c}{19.5} \\

\textsc{GenSF + \textsc{lad}}  &  \multicolumn{1}{c}{\textbf{50.9}} \\

  \cmidrule(lr){1-2}  \multicolumn{2}{c}{\textbf{Non Zero-Shot Results}} \\   \cmidrule(lr){1-2}
\textsc{GenSF (64 utterances)} & \multicolumn{1}{c}{72.2} \\
\textsc{GenSF (8633 utterances)} & \multicolumn{1}{c}{96.1} \\
\bottomrule
\end{tabularx}}%
\caption{Experimental results on the Restaurant8k corpus. We compare \textsc{GenSF + \textsc{lad}} with zero-shot results reported by prior work. For reference, we also show the performance of models (reported by prior work) when trained in few-shot and full-shot settings. }
\label{tab:slot}
\end{table}

For slot filling, experiments are carried out with \textsc{GenSF} \citep{mehri2021gensf} which currently has SoTA results on the Restaurant8k corpus \citep{coope2020span}, in both zero-shot and full-shot settings. \textsc{GenSF} reformulates slot filling as response generation in order to better leverage the capabilities of DialoGPT \citep{zhang2019dialogpt}. 

As shown in Table \ref{tab:slot}, \textsc{GenSF + \textsc{lad}} achieves a \textbf{+31.4} F-1 improvement over \textsc{GenSF} on the test set of Restaurant8k, without observing any examples from the corpus. \textsc{GenSF + \textsc{lad}} learns to detect slots in the restaurant domain given only the schema, which consists of (1) a single manually written utterance for each slot type and (2) a collection of up to 20 slot values for each slot type. This significant performance improvement in zero-shot generalization validates the claim of this paper for the problem of slot filling. \textsc{lad} is able to create synthetic data which effectively teaches \textsc{GenSF} the necessary structural constraints.

However, as shown by \citet{mehri2021gensf}, \textsc{GenSF} achieves a 72.2 F-1 score by only observing 64 human-written examples. Despite the relative success of \textsc{lad} in zero-shot settings, there remains significant room for improvement.

\begin{table}[!t]
\centering
\renewcommand*{\arraystretch}{1.1}
{
\begin{tabularx}{\linewidth}{l X }
\toprule
  \textbf{Model} & \multicolumn{1}{c}{\bf \textsc{F-1}} \\
  
  &  \\[-2ex] 
 
  \cmidrule(lr){1-2}  \multicolumn{2}{c}{\textbf{Zero-Shot Results}} \\   \cmidrule(lr){1-2}

\textsc{BERT+S {\small \citep{mosig2020star}}}  &  \multicolumn{1}{c}{28.12} \\

\textsc{SAM {\small \citep{mehri2021schema}}}  &  \multicolumn{1}{c}{53.31} \\

\textsc{SAM + \textsc{lad}}  &  \multicolumn{1}{c}{\textbf{64.36}} \\

  \cmidrule(lr){1-2}  \multicolumn{2}{c}{\textbf{Full-Shot Results}} \\   \cmidrule(lr){1-2}
\textsc{SAM {\small \citep{mehri2021schema}}}  &  \multicolumn{1}{c}{70.38} \\ \bottomrule
\end{tabularx}}%
\caption{Experimental results on the STAR corpus.  \textsc{SAM + lad} is compared with zero-shot results reported by prior work. For reference, the performance of \textsc{SAM} when trained on the full corpus is also shown. }
\label{tab:nap}
\end{table}

\begin{table*}[!t]
\centering
\renewcommand*{\arraystretch}{1.1}
{
\begin{tabularx}{\linewidth}{l X  X  X}
\toprule
  \textbf{Model (Training Data)} & \multicolumn{1}{c}{\bf \textsc{Complete \%}} & \multicolumn{1}{c}{\bf \textsc{Asks All \%}} & \multicolumn{1}{c}{\bf \textsc{Avoids Redundancy \%}} \\
  \cmidrule(lr){1-1} \cmidrule(lr){2-2} \cmidrule(lr){3-3} \cmidrule(lr){4-4}

 &    & &  \\[-2ex] 
 
\textsc{SAM (zero-shot)}  & \multicolumn{1}{c}{98.02}   & \multicolumn{1}{c}{76.15}  & \multicolumn{1}{c}{78.90} \\

\textsc{SAM (full-shot)}  & \multicolumn{1}{c}{98.31}   & \multicolumn{1}{c}{75.69}  & \multicolumn{1}{c}{80.65} \\

\textsc{SAM + lad}  & \multicolumn{1}{c}{98.52}   & \multicolumn{1}{c}{\textbf{78.39}}  & \multicolumn{1}{c}{79.13} \\ \bottomrule
\end{tabularx}}%
\caption{Results of the interactive human evaluation. We compare three models: (1) \textsc{SAM (zero-shot)}, (2) \textsc{SAM (full-shot)} and (3) \textsc{SAM + lad}. The three columns correspond to the three post-dialog questions: (1) task completion, (2) asking all necessary information and (3) avoiding redundancy. Results in boldface are statistically significant by one-tailed t-test ($p < 0.05$).}
\label{tab:human}
\end{table*}

\subsection{Next Action Prediction}

Next action prediction is particularly challenging due to the complexity of the structural constraints. In addition to the constraints of intent prediction and slot filling, next action prediction models must also learn to follow the \textit{dialog policy}. \textsc{SAM} \citep{mehri2021schema} learns to predict the system action by attending to a graph-based representation of the dialog policy. Explicitly attending to the dialog policy is an inductive bias that facilitates zero-shot generalization to unseen tasks.

Table \ref{tab:nap} shows the results for three models. \textsc{BERT+S} \citep{mosig2020star} trains a BERT model to attend to a rudimentary graph-based representation of the dialog policy. \textsc{SAM} \citep{mehri2021schema} improves the model architecture and introduces more expressive policy graphs. These two models are trained on the STAR corpus, which includes 24 different tasks and 24 different policy graphs. The zero-shot results are obtained by training on $n-1$ tasks (i.e., 23) and evaluating on the remaining task, repeated 24 times. In contrast, \textsc{SAM + lad} observes \textbf{no human-written dialogs} whatsoever. Instead, \textsc{SAM+lad} is trained only on the synthetic dialogs produced by \textsc{lad}.

In the zero-shot setting, \textsc{SAM + \textsc{lad}} achieves an \textbf{+11.05} F-1 improvement over \textsc{SAM}. Furthermore, this result is only \textbf{6.02} points below the full-shot results of \textsc{SAM}. This significant gain further validates the claim of this paper. \textsc{SAM + lad} learns the necessary structural constraints using only the synthetic data produced by \textsc{lad}.




\subsection{Interactive Human Evaluation}


\textsc{SAM + lad} achieves strong zero-shot results on the STAR corpus, especially relative to the performance of \textsc{SAM (full-shot)}. This leads us to question the performance gap between these two models. Is the full-shot model better at next action prediction, or is it just better at modelling artifacts in the STAR corpus? STAR is known to have some degree of inconsistency with the policy graphs \citep{mosig2020star}. Furthermore, static evaluation is not necessarily reflective of the performance of a model in \textbf{real settings}. Because of variable user behavior, there may be a distribution shift between the STAR corpus and interactive settings. To this end, we perform an interactive human evaluation using Amazon Mechanical Turk (AMT).

Three models are evaluated: (1) \textsc{SAM (zero-shot)}, (2) \textsc{SAM (full-shot)} and (3) \textsc{SAM + lad}. Ten scenarios are defined, each of which consists of an objective (e.g, \textit{`You want to plan a party'}) and slot values (e.g., \texttt{Name:} Kevin, \texttt{Date:} Sunday, \texttt{Num Guests:} 85). An AMT worker is instructed to interact with a dialog system according to the provided scenario. Upon completion of the dialog, three questions are answered:

\begin{enumerate}
    \item[1.] \small{Did the system successfully complete the dialog?}
    \item[2.] Did the system ask for all of the necessary information?
    \item[3.] Did the system ask for information that you had already provided it? 
\end{enumerate}

The instructions (see Appendix) tell the worker to interact \textit{naturally} (e.g., by providing information out of turn). Detailed instructions, including examples and counter-examples, are provided for the three post-dialog questions. Pre-screening is performed to ensure that AMT workers read and understood these instructions. During pre-screening, the worker must answer the post-dialog questions given two completed dialogs and the corresponding scenarios. Workers with a score of at least 5/6 qualify to participate in the interactive evaluation (45\% of workers pass the pre-screening). Pre-screening is paid \$0.75USD, regardless of the result. Each HIT (Human Intelligence Task) of the interactive evaluation includes five scenarios and pays \$3.25USD (approx. 10 minutes). A post-hoc quality check is performed to remove erroneous annotations. Simple heuristics are constructed to predict the post-dialog answers and any discrepancies with the annotations are manually verified. If an error is identified through manual validation, the annotation is removed. This form of validation is a necessary alternative to outlier detection or measures of inter-annotator agreement, since interactive dialogs are independent thereby making standard measures of data quality unsuitable.

1628 dialogs were collected, with at least 500 for each system. The results, shown in Table \ref{tab:human}, demonstrate that the performance of all three models is fairly similar in interactive settings. For the second post-dialog question, \textsc{SAM+lad} asks for all of the necessary slots \textbf{+2.7\%} more often. Assuming that the number of observations is equal to the total number of turns, this result is statistically significant ($p < 0.05$) by one-tailed t-test. 

Both \textsc{SAM (full-shot)} and \textsc{SAM (zero-shot)} are trained on human dialogs, though the latter does not observe data from the target task. In contrast, \textsc{SAM + lad} is trained only on synthetic data produced by \textsc{lad}. The comparable performance of \textsc{SAM (zero-shot)} and \textsc{SAM (full-shot)} is noteworthy and can potentially be explained by two facts: (1) the interactive dialogs are sampled from a different distribution (e.g., more informal, typos, more slots per utterance) from the STAR corpus, making the evaluation equally difficult for both systems, (2) \textsc{SAM (zero-shot)} has observed dialogs from the domain (e.g., seen \texttt{bank-balance} when evaluating on \texttt{bank-fraud-report}). Despite not observing any human dialogs, in interactive settings \textsc{SAM + lad} attains zero-shot performance comparable to training on human dialogs from the STAR corpus. Though there remains significant room for improvement, the results of this interactive human evaluation demonstrate the efficacy of \textsc{lad}. By leveraging the strengths of large language models to induce linguistic diversity, \textsc{lad} produces synthetic data that effectively conveys the necessary structural constraints and facilitates zero-shot generalization, even in challenging interactive settings.


\section{Related Work} 

\subsection{User Simulators for Task-Oriented Dialog}

The use of synthetic data in task-oriented dialog is a long-standing approach. Early dialog research leveraged user simulators for evaluation and optimization \citep{eckert1997user,scheffler2000probabilistic,schatzmann2006survey}. \citet{schatzmann2007agenda} propose a probabilistic agenda-based user simulator for bootstrapping a POMDB dialog system, demonstrating reasonable task completion rates. \citet{georgila2006user} train an n-gram user simulator which models both ASR and understanding errors. \citet{gonzalez2010cooperative} explicitly model user cooperativeness in a statistical user simulator. 

\citet{li2016user} propose an agenda-based user simulator for training dialog policies with RL. \citet{crook2017sequence} train a sequence-to-sequence model for user simulation. \citet{kreyssig2018neural} introduce the neural user simulator (NUS), which trains a sequence-to-sequence network conditioned on user goals and the dialog history, outperforming existing methods on an interactive evaluation. \citet{shi2019build} carry out a comprehensive analysis of six different user simulators, with different dialog planning and generation methods. A key takeaway of this analysis is using agenda-based simulators to train RL systems generally results in higher performance on human evaluation. \citet{lin2021domain} propose a domain-indepdendent transformer-based user simulator (TUS). The feature representations of TUS are domain-independent, thereby facilitating learning of cross-domain user behavior. TUS is trained on MultiWOZ \citep{budzianowski2018multiwoz} and can effectively transfer to unseen domains.

\textsc{lad} can be characterized as an agenda-based simulator, wherein the schema describes the ontology and the policy. The core novelty of \textsc{lad} in the context of prior work is three-fold: (1) large LMs to induce linguistic diversity, (2) \textit{zero-shot} domain-agnostic synthetic data creation, and (3) the schema as a standardized expression of structural constraints. \textsc{lad} can potentially be further improved by incorporating strategies from prior work, such as modelling cooperativeness \citep{gonzalez2010cooperative} or ASR errors \citep{georgila2006user}.













\subsection{Using Large Language Models}

Large language models \citep{brown2020language,chowdhery2022palm} exhibit strong language understanding, generation and reasoning abilities. Prompting is the dominant paradigm for leveraging large LMs for various downstream problems \citep{reynolds2021prompt,lester2021power}. \citet{madotto2021few} demonstrate the efficacy of few-shot prompting for both open-domain and task-oriented dialog, with a focus on response generation and conversational parsing.

Several papers have used GPT-3 to generate synthetic data \citep{yoo2021gpt3mix,wang2021towards}. These approaches rely on GPT-3 to generate the labels and are not suitable for task-oriented dialog.  To our knowledge, \textsc{lad} is the first paper to leverage large LMs to \textit{reformulate} utterances, in order to create synthetic data for task-oriented dialog.

\section{Conclusion} 

In an effort to leverage the abilities of large LMs to facilitate zero-shot generalization in task-oriented dialog, this paper introduces \textsc{\textbf{lad}}. \textsc{lad} creates diverse and accurate synthetic data, in order to convey the necessary setting-specific structural constraints to neural dialog models. \textsc{lad} achieves significant performance gains on zero-shot intent prediction, slot filling and next action prediction across five datasets. Furthermore, \textsc{lad} is shown to perform competitively in interactive human evaluation, without observing human-annotated data. 

\section{Acknowledgements}

This work was funded by a Google Research Collabs grant.

\bibliography{anthology,custom}

\begin{thebibliography}{40}
\expandafter\ifx\csname natexlab\endcsname\relax\def\natexlab#1{#1}\fi

\bibitem[{Brown et~al.(2020)Brown, Mann, Ryder, Subbiah, Kaplan, Dhariwal,
  Neelakantan, Shyam, Sastry, Askell et~al.}]{brown2020language}
Tom Brown, Benjamin Mann, Nick Ryder, Melanie Subbiah, Jared~D Kaplan, Prafulla
  Dhariwal, Arvind Neelakantan, Pranav Shyam, Girish Sastry, Amanda Askell,
  et~al. 2020.
\newblock Language models are few-shot learners.
\newblock \emph{Advances in neural information processing systems},
  33:1877--1901.

\bibitem[{Budzianowski et~al.(2018)Budzianowski, Wen, Tseng, Casanueva, Ultes,
  Ramadan, and Ga{\v{s}}i{\'c}}]{budzianowski2018multiwoz}
Pawe{\l} Budzianowski, Tsung-Hsien Wen, Bo-Hsiang Tseng, Inigo Casanueva,
  Stefan Ultes, Osman Ramadan, and Milica Ga{\v{s}}i{\'c}. 2018.
\newblock Multiwoz--a large-scale multi-domain wizard-of-oz dataset for
  task-oriented dialogue modelling.
\newblock \emph{arXiv preprint arXiv:1810.00278}.

\bibitem[{Casanueva et~al.(2020)Casanueva, Tem{\v{c}}inas, Gerz, Henderson, and
  Vuli{\'c}}]{casanueva2020efficient}
Inigo Casanueva, Tadas Tem{\v{c}}inas, Daniela Gerz, Matthew Henderson, and
  Ivan Vuli{\'c}. 2020.
\newblock Efficient intent detection with dual sentence encoders.
\newblock \emph{arXiv preprint arXiv:2003.04807}.

\bibitem[{Chowdhery et~al.(2022)Chowdhery, Narang, Devlin, Bosma, Mishra,
  Roberts, Barham, Chung, Sutton, Gehrmann et~al.}]{chowdhery2022palm}
Aakanksha Chowdhery, Sharan Narang, Jacob Devlin, Maarten Bosma, Gaurav Mishra,
  Adam Roberts, Paul Barham, Hyung~Won Chung, Charles Sutton, Sebastian
  Gehrmann, et~al. 2022.
\newblock Palm: Scaling language modeling with pathways.
\newblock \emph{arXiv preprint arXiv:2204.02311}.

\bibitem[{Coope et~al.(2020)Coope, Farghly, Gerz, Vuli{\'c}, and
  Henderson}]{coope2020span}
Sam Coope, Tyler Farghly, Daniela Gerz, Ivan Vuli{\'c}, and Matthew Henderson.
  2020.
\newblock Span-convert: Few-shot span extraction for dialog with pretrained
  conversational representations.
\newblock \emph{arXiv preprint arXiv:2005.08866}.

\bibitem[{Crook and Marin(2017)}]{crook2017sequence}
Paul~A Crook and Alex Marin. 2017.
\newblock Sequence to sequence modeling for user simulation in dialog systems.
\newblock In \emph{INTERSPEECH}, pages 1706--1710.

\bibitem[{Devlin et~al.(2018)Devlin, Chang, Lee, and
  Toutanova}]{devlin2018bert}
Jacob Devlin, Ming-Wei Chang, Kenton Lee, and Kristina Toutanova. 2018.
\newblock Bert: Pre-training of deep bidirectional transformers for language
  understanding.
\newblock \emph{arXiv preprint arXiv:1810.04805}.

\bibitem[{Eckert et~al.(1997)Eckert, Levin, and Pieraccini}]{eckert1997user}
Wieland Eckert, Esther Levin, and Roberto Pieraccini. 1997.
\newblock User modeling for spoken dialogue system evaluation.
\newblock In \emph{1997 IEEE Workshop on Automatic Speech Recognition and
  Understanding Proceedings}, pages 80--87. IEEE.

\bibitem[{Feng et~al.(2021)Feng, Gangal, Wei, Chandar, Vosoughi, Mitamura, and
  Hovy}]{feng2021survey}
Steven~Y Feng, Varun Gangal, Jason Wei, Sarath Chandar, Soroush Vosoughi,
  Teruko Mitamura, and Eduard Hovy. 2021.
\newblock A survey of data augmentation approaches for nlp.
\newblock \emph{arXiv preprint arXiv:2105.03075}.

\bibitem[{Georgila et~al.(2006)Georgila, Henderson, and
  Lemon}]{georgila2006user}
Kallirroi Georgila, James Henderson, and Oliver Lemon. 2006.
\newblock User simulation for spoken dialogue systems: learning and evaluation.
\newblock In \emph{Interspeech}, pages 1065--1068. Citeseer.

\bibitem[{Gonz{\'a}lez et~al.(2010)Gonz{\'a}lez, Quarteroni, Riccardi, and
  Varges}]{gonzalez2010cooperative}
Meritxell Gonz{\'a}lez, Silvia Quarteroni, Giuseppe Riccardi, and Sebastian
  Varges. 2010.
\newblock Cooperative user models in statistical dialog simulators.
\newblock In \emph{Proceedings of the SIGDIAL 2010 Conference}, pages 217--220.

\bibitem[{Harris(1954)}]{harris1954distributional}
Zellig~S Harris. 1954.
\newblock Distributional structure.
\newblock \emph{Word}, 10(2-3):146--162.

\bibitem[{Henderson and Vuli{\'c}(2020)}]{henderson2020convex}
Matthew Henderson and Ivan Vuli{\'c}. 2020.
\newblock Convex: Data-efficient and few-shot slot labeling.
\newblock \emph{arXiv preprint arXiv:2010.11791}.

\bibitem[{Kreyssig et~al.(2018)Kreyssig, Casanueva, Budzianowski, and
  Gasic}]{kreyssig2018neural}
Florian Kreyssig, Inigo Casanueva, Pawel Budzianowski, and Milica Gasic. 2018.
\newblock Neural user simulation for corpus-based policy optimisation for
  spoken dialogue systems.
\newblock \emph{arXiv preprint arXiv:1805.06966}.

\bibitem[{Larson et~al.(2019)Larson, Mahendran, Peper, Clarke, Lee, Hill,
  Kummerfeld, Leach, Laurenzano, Tang, and Mars}]{larson-etal-2019-evaluation}
Stefan Larson, Anish Mahendran, Joseph~J. Peper, Christopher Clarke, Andrew
  Lee, Parker Hill, Jonathan~K. Kummerfeld, Kevin Leach, Michael~A. Laurenzano,
  Lingjia Tang, and Jason Mars. 2019.
\newblock \href {https://doi.org/10.18653/v1/D19-1131} {An evaluation dataset
  for intent classification and out-of-scope prediction}.
\newblock In \emph{Proceedings of the 2019 Conference on Empirical Methods in
  Natural Language Processing and the 9th International Joint Conference on
  Natural Language Processing (EMNLP-IJCNLP)}, pages 1311--1316, Hong Kong,
  China. Association for Computational Linguistics.

\bibitem[{Lester et~al.(2021)Lester, Al-Rfou, and Constant}]{lester2021power}
Brian Lester, Rami Al-Rfou, and Noah Constant. 2021.
\newblock The power of scale for parameter-efficient prompt tuning.
\newblock \emph{arXiv preprint arXiv:2104.08691}.

\bibitem[{Li et~al.(2016)Li, Lipton, Dhingra, Li, Gao, and Chen}]{li2016user}
Xiujun Li, Zachary~C Lipton, Bhuwan Dhingra, Lihong Li, Jianfeng Gao, and
  Yun-Nung Chen. 2016.
\newblock A user simulator for task-completion dialogues.
\newblock \emph{arXiv preprint arXiv:1612.05688}.

\bibitem[{Lin et~al.(2021)Lin, Lubis, Hu, van Niekerk, Geishauser, Heck, Feng,
  and Ga{\v{s}}i{\'c}}]{lin2021domain}
Hsien-chin Lin, Nurul Lubis, Songbo Hu, Carel van Niekerk, Christian
  Geishauser, Michael Heck, Shutong Feng, and Milica Ga{\v{s}}i{\'c}. 2021.
\newblock Domain-independent user simulation with transformers for
  task-oriented dialogue systems.
\newblock \emph{arXiv preprint arXiv:2106.08838}.

\bibitem[{Liu et~al.(2019{\natexlab{a}})Liu, Eshghi, Swietojanski, and
  Rieser}]{liu2019benchmarking}
Xingkun Liu, Arash Eshghi, Pawel Swietojanski, and Verena Rieser.
  2019{\natexlab{a}}.
\newblock Benchmarking natural language understanding services for building
  conversational agents.
\newblock \emph{arXiv preprint arXiv:1903.05566}.

\bibitem[{Liu et~al.(2019{\natexlab{b}})Liu, Ott, Goyal, Du, Joshi, Chen, Levy,
  Lewis, Zettlemoyer, and Stoyanov}]{liu2019roberta}
Yinhan Liu, Myle Ott, Naman Goyal, Jingfei Du, Mandar Joshi, Danqi Chen, Omer
  Levy, Mike Lewis, Luke Zettlemoyer, and Veselin Stoyanov. 2019{\natexlab{b}}.
\newblock Roberta: A robustly optimized bert pretraining approach.
\newblock \emph{arXiv preprint arXiv:1907.11692}.

\bibitem[{Liu et~al.(2020)Liu, Winata, Xu, and Fung}]{liu2020coach}
Zihan Liu, Genta~Indra Winata, Peng Xu, and Pascale Fung. 2020.
\newblock Coach: A coarse-to-fine approach for cross-domain slot filling.
\newblock \emph{arXiv preprint arXiv:2004.11727}.

\bibitem[{Madotto et~al.(2021)Madotto, Lin, Winata, and Fung}]{madotto2021few}
Andrea Madotto, Zhaojiang Lin, Genta~Indra Winata, and Pascale Fung. 2021.
\newblock Few-shot bot: Prompt-based learning for dialogue systems.
\newblock \emph{arXiv preprint arXiv:2110.08118}.

\bibitem[{Mehri and Eric(2021)}]{mehri2021example}
Shikib Mehri and Mihail Eric. 2021.
\newblock Example-driven intent prediction with observers.
\newblock In \emph{Proceedings of the 2021 Conference of the North American
  Chapter of the Association for Computational Linguistics: Human Language
  Technologies}, pages 2979--2992.

\bibitem[{Mehri et~al.(2020)Mehri, Eric, and Hakkani-Tur}]{mehri2020dialoglue}
Shikib Mehri, Mihail Eric, and Dilek Hakkani-Tur. 2020.
\newblock Dialoglue: A natural language understanding benchmark for
  task-oriented dialogue.
\newblock \emph{arXiv preprint arXiv:2009.13570}.

\bibitem[{Mehri and Eskenazi(2021{\natexlab{a}})}]{mehri2021gensf}
Shikib Mehri and Maxine Eskenazi. 2021{\natexlab{a}}.
\newblock Gensf: Simultaneous adaptation of generative pre-trained models and
  slot filling.
\newblock In \emph{Proceedings of the 22nd Annual Meeting of the Special
  Interest Group on Discourse and Dialogue}, pages 489--498.

\bibitem[{Mehri and Eskenazi(2021{\natexlab{b}})}]{mehri2021schema}
Shikib Mehri and Maxine Eskenazi. 2021{\natexlab{b}}.
\newblock Schema-guided paradigm for zero-shot dialog.
\newblock In \emph{Proceedings of the 22nd Annual Meeting of the Special
  Interest Group on Discourse and Dialogue}, pages 499--508.

\bibitem[{Mitchell(1980)}]{mitchell1980need}
Tom~M Mitchell. 1980.
\newblock \emph{The need for biases in learning generalizations}.
\newblock Department of Computer Science, Laboratory for Computer Science
  Research~….

\bibitem[{Mosig et~al.(2020)Mosig, Mehri, and Kober}]{mosig2020star}
Johannes~EM Mosig, Shikib Mehri, and Thomas Kober. 2020.
\newblock Star: A schema-guided dialog dataset for transfer learning.
\newblock \emph{arXiv preprint arXiv:2010.11853}.

\bibitem[{Peng et~al.(2020)Peng, Li, Li, Shayandeh, Liden, and
  Gao}]{peng2020soloist}
Baolin Peng, Chunyuan Li, Jinchao Li, Shahin Shayandeh, Lars Liden, and
  Jianfeng Gao. 2020.
\newblock Soloist: Few-shot task-oriented dialog with a single pretrained
  auto-regressive model.
\newblock \emph{arXiv preprint arXiv:2005.05298}.

\bibitem[{Rastogi et~al.(2020)Rastogi, Zang, Sunkara, Gupta, and
  Khaitan}]{rastogi2020towards}
Abhinav Rastogi, Xiaoxue Zang, Srinivas Sunkara, Raghav Gupta, and Pranav
  Khaitan. 2020.
\newblock Towards scalable multi-domain conversational agents: The
  schema-guided dialogue dataset.
\newblock In \emph{Proceedings of the AAAI Conference on Artificial
  Intelligence}, volume~34, pages 8689--8696.

\bibitem[{Reynolds and McDonell(2021)}]{reynolds2021prompt}
Laria Reynolds and Kyle McDonell. 2021.
\newblock Prompt programming for large language models: Beyond the few-shot
  paradigm.
\newblock In \emph{Extended Abstracts of the 2021 CHI Conference on Human
  Factors in Computing Systems}, pages 1--7.

\bibitem[{Schatzmann et~al.(2007)Schatzmann, Thomson, Weilhammer, Ye, and
  Young}]{schatzmann2007agenda}
Jost Schatzmann, Blaise Thomson, Karl Weilhammer, Hui Ye, and Steve Young.
  2007.
\newblock Agenda-based user simulation for bootstrapping a pomdp dialogue
  system.
\newblock In \emph{Human Language Technologies 2007: The Conference of the
  North American Chapter of the Association for Computational Linguistics;
  Companion Volume, Short Papers}, pages 149--152.

\bibitem[{Schatzmann et~al.(2006)Schatzmann, Weilhammer, Stuttle, and
  Young}]{schatzmann2006survey}
Jost Schatzmann, Karl Weilhammer, Matt Stuttle, and Steve Young. 2006.
\newblock A survey of statistical user simulation techniques for
  reinforcement-learning of dialogue management strategies.
\newblock \emph{The knowledge engineering review}, 21(2):97--126.

\bibitem[{Scheffler and Young(2000)}]{scheffler2000probabilistic}
Konrad Scheffler and Steve Young. 2000.
\newblock Probabilistic simulation of human-machine dialogues.
\newblock In \emph{2000 IEEE International Conference on Acoustics, Speech, and
  Signal Processing. Proceedings (Cat. No. 00CH37100)}, volume~2, pages
  II1217--II1220. IEEE.

\bibitem[{Shi et~al.(2019)Shi, Qian, Wang, and Yu}]{shi2019build}
Weiyan Shi, Kun Qian, Xuewei Wang, and Zhou Yu. 2019.
\newblock How to build user simulators to train rl-based dialog systems.
\newblock \emph{arXiv preprint arXiv:1909.01388}.

\bibitem[{Sinha et~al.(2021)Sinha, Jia, Hupkes, Pineau, Williams, and
  Kiela}]{sinha2021masked}
Koustuv Sinha, Robin Jia, Dieuwke Hupkes, Joelle Pineau, Adina Williams, and
  Douwe Kiela. 2021.
\newblock Masked language modeling and the distributional hypothesis: Order
  word matters pre-training for little.
\newblock \emph{arXiv preprint arXiv:2104.06644}.

\bibitem[{Wang et~al.(2021)Wang, Yu, Firat, and Cao}]{wang2021towards}
Zirui Wang, Adams~Wei Yu, Orhan Firat, and Yuan Cao. 2021.
\newblock Towards zero-label language learning.
\newblock \emph{arXiv preprint arXiv:2109.09193}.

\bibitem[{Wei et~al.(2021)Wei, Bosma, Zhao, Guu, Yu, Lester, Du, Dai, and
  Le}]{wei2021finetuned}
Jason Wei, Maarten Bosma, Vincent~Y Zhao, Kelvin Guu, Adams~Wei Yu, Brian
  Lester, Nan Du, Andrew~M Dai, and Quoc~V Le. 2021.
\newblock Finetuned language models are zero-shot learners.
\newblock \emph{arXiv preprint arXiv:2109.01652}.

\bibitem[{Yoo et~al.(2021)Yoo, Park, Kang, Lee, and Park}]{yoo2021gpt3mix}
Kang~Min Yoo, Dongju Park, Jaewook Kang, Sang-Woo Lee, and Woomyeong Park.
  2021.
\newblock Gpt3mix: Leveraging large-scale language models for text
  augmentation.
\newblock \emph{arXiv preprint arXiv:2104.08826}.

\bibitem[{Zhang et~al.(2019)Zhang, Sun, Galley, Chen, Brockett, Gao, Gao, Liu,
  and Dolan}]{zhang2019dialogpt}
Yizhe Zhang, Siqi Sun, Michel Galley, Yen-Chun Chen, Chris Brockett, Xiang Gao,
  Jianfeng Gao, Jingjing Liu, and Bill Dolan. 2019.
\newblock Dialogpt: Large-scale generative pre-training for conversational
  response generation.
\newblock \emph{arXiv preprint arXiv:1911.00536}.

\end{thebibliography}
\bibliographystyle{acl_natbib}

\appendix

\newpage
\section{Appendix}
\label{sec:appendix}

\subsection{Algorithm for Next Action Prediction}

Algorithm \ref{alg:cap} describes the probabilistic process of creating seed dialogs for next action prediction, conditioned on the schema. The input to this algorithm is the schema, which consists of (1) an utterance for each intent, (2) a set of slot values and (3) a graph-based representation of the dialog policy. A single dialog is produced, with variable user behavior and deterministic system behavior.

\begin{algorithm}
\caption{Next action prediction seed data generation. This algorithm defines the process of generating a \textit{single} dialog from the schema. The input to this function is the schema, consisting of a set of utterances, slot values and the graph-based representation of the dialog policy.}\label{alg:cap}
\begin{algorithmic}
\Function{GenDialog}{$\mathcal{U}, \mathcal{S}, \mathcal{G}_{policy}$ }
\State $cur \gets \mathrm{start\_node}(\mathcal{G}_{policy})$
\State $\mathcal{D} \gets 
\{\}$
\While{$cur \neq \mathrm{end\_node}(\mathcal{G}_{policy})$}

\State $r \gets \mathrm{random}(0, 1)$
\State $u_{cur} \gets \mathrm{utt}(\mathcal{U}, cur)$
\If{$r < 0.15$} 
  \LeftComment{Combine three utterances.}
  \State $n_2 \gets \mathrm{sample}(\mathrm{unvisited}(\mathcal{G}_{policy}))$
  \State $n_3 \gets \mathrm{sample}(\mathrm{unvisited}(\mathcal{G}_{policy}))$
  \State $u_2 \gets \mathrm{utt}(\mathcal{U}, n_2)$
  \State $u_3 \gets \mathrm{utt}(\mathcal{U}, n_3)$
  \State $u_{cur} \gets \mathrm{concat}(u_{cur}, u_2, u_3)$
\ElsIf{$r < 0.5$}
  \LeftComment{Combine two utterances.}
  \State $n_2 \gets \mathrm{sample}(\mathrm{unvisited}(\mathcal{G}_{policy}))$
  \State $u_2 \gets \mathrm{utt}(\mathcal{U}, n_2)$
  \State $u_{cur} \gets \mathrm{concat}(u_{cur}, u_2)$
\Else
  \LeftComment{Use the single utterance.}
  \State $u_{cur} \gets u_{cur}$
\EndIf \\

\LeftComment{Sample slot values. Add $u_{cur}$ to dialog.}
\State $u_{cur} \gets \mathrm{sample\_slots}(\mathcal{S}, u_{cur})$
\State $\mathcal{D} \gets \mathcal{D} + u_{cur}$ \\

\LeftComment{Add system action according to policy.}
\State $n_{sys} \gets \mathrm{policy}(\mathcal{D}, \mathcal{G}_{policy})$ 
\State $u_{sys} \gets \mathrm{utt}(\mathcal{U}, n_{sys})$
\State $\mathcal{D} \gets \mathcal{D} + u_{sys}$\\

\LeftComment{Sample next user node using the policy.}
\State $cur \gets \mathrm{sample}(\mathrm{next}(\mathcal{D}, \mathcal{G}_{policy}))$
\EndWhile

\State \Return $\mathcal{D}$
\EndFunction

\end{algorithmic}
\end{algorithm}

\subsection{GPT-3 Details}

The hyperparameters for GPT-3 are shown in Table \ref{tab:hp}. The prompt used to induce linguistic diversity is shown in Table \ref{tab:prompt}. Both the prompt and the hyperparameters were identified through manual experimentation, rather than any exhaustive search.

This specific prompt was chosen for two reasons. First, it specifically instructs GPT-3 (through the instruction and examples) to generate \textit{five} different reformulations. As such, the in-context examples influence GPT-3 to produce \textit{diverse} reformulations. The linguistic diversity is derived both from the sampling in the decoding algorithm, and from the prompt. Second, this prompt explicitly instructs GPT-3 to reformulate the concatenation of disjoint utterances into a single natural utterance. Since Algorithm \ref{alg:cap} concatenates template utterances, this facilitates natural reformulations. While we avoid making claims about our prompt being optimal, these two properties make it suitable for our objective.

Future work should explore better prompts. One potential direction is to explicitly enforce that certain slot values be present in the reformulated utterance. This would ensure that the reformulated utterances adhere to the schema, and remove the need for the validation step of \textsc{lad}. 

\begin{table}[h]
\centering
\renewcommand*{\arraystretch}{1.1}
{
\begin{tabularx}{\linewidth}{l X }
\toprule
  \textbf{Parameter} & \multicolumn{1}{c}{\textbf{Value}} \\
  
  &  \\[-2ex] 
 

 \texttt{engine}  &  \multicolumn{1}{c}{ \texttt{davinci}} \\
 \texttt{temperature}  &  \multicolumn{1}{c}{ \texttt{0.7}} \\
 \texttt{max\_tokens}  &  \multicolumn{1}{c}{ \texttt{250}} \\
 \texttt{frequency\_penalty}  &  \multicolumn{1}{c}{ \texttt{0.22}} \\
 \texttt{presence\_penalty}  &  \multicolumn{1}{c}{ \texttt{0}} \\
 \texttt{stop}  &  \multicolumn{1}{c}{ \texttt{==, Original}} \\
\bottomrule
\end{tabularx}}%
\caption{GPT-3 hyperparameters.}
\label{tab:hp}
\end{table}

\begin{table*}
\centering
\renewcommand*{\arraystretch}{1.1}
{
\begin{tabularx}{\linewidth}{p{\linewidth}}
\toprule
  \textbf{GPT-3 Prompt} \\
 Given a set of sentences, generate 5 natural utterances that convey the same meaning. \\
== \\
Original: My name is John. I want to reserve a flight. \\
1. The name is John and I need to reserve a flight. \\
2. I'm John and I'd like to book a flight. \\
3. I'm booking a flight and my name is John. \\
4. I want to reserve a flight, under the name John. \\
5. I am John and I'd like to make a flight reservation. \\
== \\
Original: The name of the hotel is Shadyside Inn. I will be arriving May 7th. I will be leaving May 27th. \\
1. I want a reservation for the Shadyside Inn from May 7th to the 27th. \\
2. Please reserve the Shadyside Inn from the 7th to the 27th of May. \\
3. I want you to reserve the Shadyside Inn from May 7th to the 27th. \\
4. I'll be in town from May 7th to the 27th and I'd like to stay at the Shadyside Inn during that time. \\
5. Give me a room at the Shadyside Inn from  May 7th to the 27th. \\
== \\
Original: I would like to follow-up on my appointment with Dr. Veronica. My name is Andy. My appointment was on Friday. \\
1. I'm Andy and I had an appointment with Dr. Veronica on Friday. I'd like to follow up with her. \\
2. My name is Andy, I'd like to follow up on my appointment with Dr. Veronica on Friday. \\
3. I had an appointment with Dr. Veronica last Friday and I'd like to follow up. My name is Andy. \\
4. Andy here. I'm trying to follow-up on my appointment with Dr. Veronica last Friday. \\
5. Can you help me follow-up on an appointment with Dr. Veronica last Friday? My name is Andy. \\
== \\
Original: Thanks \\
1. Thank you! \\
2. Thanks for your help. \\
3. Thank you so much. \\
4. Thank you for helping \\
5. thx \\
== \\
Original:  The meeting should be on Monday. The reason for the meeting is to allocate department budgets. \\
1. The meeting is Monday and the reason for the meeting is to allocate department budgets. \\
2. The meeting will be held Monday to discuss department budgets. \\
3. The meeting is scheduled for Monday to discuss department budgets. \\
4. We will have the meeting on Monday to allocate department budgets. \\
5. Let's meet on Monday to allocate department budgets. \\
== \\
Original: Hello. \\
1. Hey! \\
2. hi \\
3. Hi there \\
4. Hello, can you help me? \\
5. Yo \\
\bottomrule
\end{tabularx}}%
\caption{Prompt used to induce linguistic diversity with GPT-3.}
\label{tab:prompt}
\end{table*}

\subsection{Example Data}

Tables \ref{tab:slot_lad}, \ref{tab:intent_lad}, and \ref{tab:nap_lad} show synthetic data produced by \textsc{lad} for slot filling, intent prediction and next action prediction, respectively. 

\begin{table*}
\centering
\renewcommand*{\arraystretch}{1.1}
{
\begin{tabularx}{\linewidth}{lX}
\toprule

\textbf{Original:} & My first name is Vashti. My last name is Leard. \\
\textbf{Reformulated:} & My name is Vashti Leard. \\
& My name is Vashti, last name Leard. \\
& I'm Vashti, my last name is Leard. \\
& That's Vashti Leard to the rescue! \\
& That's Vashti, last name Leard to the rescue! \\

\midrule

\textbf{Original:} & The date of my restaurant reservation is January eighteenth. My first name is Artie. The time of my restaurant reservation is 3:52 AM. \\
\textbf{Reformulated:} & The date of my restaurant reservation is January eighteenth and my first name is Artie. The time of my restaurant reservation is 3:52 AM. \\
& I have a reservation at the restaurant at 3:52 AM on January eighteenth. My name is Artie. \\
& My reservation at the restaurant is on January eighteenth and my first name is Artie. The time is 3:52 AM. \\
& I have a reservation at the restaurant on January eighteenth, at 3:52 AM and my name is Artie. \\
& I'm trying to make a reservation for January eighteenth at 3:52 AM at the restaurant and my name is Artie. \\

\midrule

\textbf{Original:} & My first name is Phyllis. I want a restaurant reservation for six adults and two kids. \\
\textbf{Reformulated:} & I want a reservation for six adults and two kids. My name is Phyllis. \\
& I need a reservation for six adults and two kids. My name is Phyllis. \\
& I'd like to reserve a table for six adults and two kids. My name is Phyllis. \\
& Could you book me a table for six adults and two kids? My name is Phyllis. \\
& I need you to book me a table for six adults and two kids. My name is Phyllis. \\

\midrule

\textbf{Original:} & My last name is Kiekbusch. I want a restaurant reservation for 4 people. The date of my restaurant reservation is October 24. \\
\textbf{Reformulated:} & I'd like to make a reservation for my group of 4 people at the restaurant on October 24th. My last name is Kiekbusch \\
& I'd like to reserve a table for my group of 4 people at the restaurant on October 24th. My last name is Kiekbusch. \\
& I want to make a reservation for 4 people at the restaurant on October 24th. My last name is Kiekbusch. \\
& Can you make a reservation for me and my group of 4 people at the restaurant on October 24th? My last name is Kiekbusch. \\ \bottomrule
\end{tabularx}}%
\caption{Examples of synthetic data produced by \textsc{lad} for slot filling on the Restaurant8k corpus \citep{coope2020span}.}
\label{tab:slot_lad}
\end{table*}

\begin{table*}
\centering
\renewcommand*{\arraystretch}{1.1}
{
\begin{tabularx}{\linewidth}{lX}
\toprule
\multicolumn{2}{c}{\textbf{\textsc{HWU64}}} \\  
\midrule

\textbf{Original:} & remind me about my alarms today (intent: \texttt{alarm\_query}) \\
\textbf{Reformulated:} & I want to know about any alarms set for today \\
& Are there any alarms today that I should know about? \\
& remind me of my alarms today \\
& Could you please fill me in on the alarms that are set for today? \\
& Do you have anything for me today? \\

\midrule

\textbf{Original:} & bed room two lights off (intent: \texttt{iot\_hue\_lightoff}) \\
\textbf{Reformulated:} & turn off bedroom two lights \\
& bedroom two lights off \\
& Lights off in bedroom two please. \\

\midrule
\multicolumn{2}{c}{\textbf{\textsc{CLINC150}}} \\  
\midrule

\textbf{Original} & do people like chipotle (intent: \texttt{restaurant\_reviews}) \\
\textbf{Reformulated} & Is chipotle good? \\
& I want to know if people like chipotle. \\
& Do people like chipotle? \\
& I'm wondering, do people like chipotle? \\
& I'd like to know what people think about chipotle. \\

\midrule

\textbf{Original:} & tell me a joke about cows (intent: \texttt{tell\_joke}) \\
\textbf{Reformulated:} & can you tell me a joke about cows? \\
& do you know any jokes about cows? \\
& was that a joke about cows? \\
& tell me a joke about cows! \\
& What do you call a cow who is hungry? \\

\midrule
\multicolumn{2}{c}{\textbf{\textsc{Banking77}}} \\  
\midrule 

\textbf{Original:} & Where can I withdraw money from? (intent: \texttt{atm\_support}) \\
\textbf{Reformulated:} & Where can I get cash from? \\
& Where can I get some cash from? \\
& What is the nearest ATM to withdraw money from? \\
& Where can I get some cash? \\
& Where is the closest ATM? \\
& Where is a cash machine? \\
\midrule
\textbf{Original:} & The app says I made a cash withdrawal that I didn't make (intent: \texttt{cash\_withdrawal\_not\_recognised}) \\
\textbf{Reformulated:} & I didn't make a cash withdrawal from this app. \\
& I didn't make a cash withdrawal from this app but it says I did. \\
& The app says I made a withdrawal but I didn't make a withdrawal. \\
& There is a cash withdrawal transaction on my account that I didn't make. \\
& My bank statement shows a cash withdrawal that I did not make. \\
& My bank indicates I made a cash withdrawal I didn't make. \\ \bottomrule
\end{tabularx}}%
\caption{Examples of synthetic data produced by \textsc{lad} for intent prediction}
\label{tab:intent_lad}
\end{table*}

\begin{table*}
\centering
\renewcommand*{\arraystretch}{1.1}
{
\begin{tabularx}{\linewidth}{p{\linewidth}}
\toprule

\textbf{User:} Good morning \\
\textbf{System:} Hello, how can I help? \\
\textbf{User:} What's the weather like in New York City? \\
\textbf{System:} For what day would you like the weather forecast? \\
\textbf{User:} It's Friday today, isn't it? \\
\textbf{System:} [QUERY] \\
\textbf{DB:} [RESULT] APIName = weather ; City = New York City ; Day = Friday ; TemperatureCelsius = 0 ; Weather = Sunny; id = 409 ; \\
\textbf{System:} It will be Sunny all day on Friday in New York City, with temperatures of around 0 degrees celsius. \\
\textbf{User:} Thanks. \\
\textbf{System:} Is there anything else that I can do for you? \\
\textbf{User:} Bye \\
\textbf{System:} Thank you and goodbye. \\

\midrule

\textbf{User:} Hello \\
\textbf{System:} Hello, how can I help? \\
\textbf{User:} I want to schedule a viewing of an apartment. I'm particularly interested in the two bedroom units. If not, the studio is also fine. \\
\textbf{System:} Could you give me your name, please? \\
\textbf{User:} My name is John and I'm a student at California State University, Northridge.\\
\textbf{System:} What apartment are you interested in?\\
\textbf{User:} Is it possible to view One on Center Apartments tonight at 2 pm?\\
\textbf{System:} What day would you like to make the booking for?\\
\textbf{User:} The viewing is going to be on Wednesday.\\
\textbf{System:} Have you already paid the application fee for the apartment?\\
\textbf{User:} The application fee is not paid and we're still waiting for it.\\
\textbf{System:} [QUERY\_CHECK]\\
\textbf{DB:} [RESULT\_CHECK] APIName = apartment\_schedule ; Message = The time slot is available. ;\\
\textbf{System:} Great, there is still a viewing available at that time. Would you like me to book it for you?\\
\textbf{User:} I'm okay with that.\\
\textbf{System:} [QUERY\_BOOK]\\
\textbf{DB}: [RESULT\_BOOK] APIName = apartment\_schedule ; Message = Your apartment viewing has been successfuly scheduled. Please bring Proof of Income and SCHUFA certificate with you. ;\\
\textbf{System:} Excellent, the booking is done for you!\\
\textbf{User:} Thanks.\\
\textbf{System:} Thank you and goodbye.\\ \bottomrule
\end{tabularx}}%
\caption{Dialogs produced by \textsc{lad} for next action prediction on the STAR corpus \citep{mosig2020star}.}
\label{tab:nap_lad}
\end{table*}

\subsection{Interactive Human Evaluation}

This section provides a detailed description of the interactive human evaluation, including (1) the annotation instructions, (2) the annotation interface, and (3) the interactive dialogs. 

\subsubsection{Annotation Instructions}

Tables \ref{tab:qual_instructions} and \ref{tab:qual_instructions2} show the instructions for the pre-screening qualification task. Figures \ref{fig:qual1} and \ref{fig:qual2} show the pre-screening examples that the AMT workers answered the three-post dialog questions for. Workers qualified to complete the annotation HIT only if their pre-screening score was at least 5/6 correct and 2/2 on the completion question (as this was the most erroneous in preliminary experiments).

\subsubsection{Annotation Interface}

\begin{figure*}
    \centering
    \includegraphics[width=\linewidth]{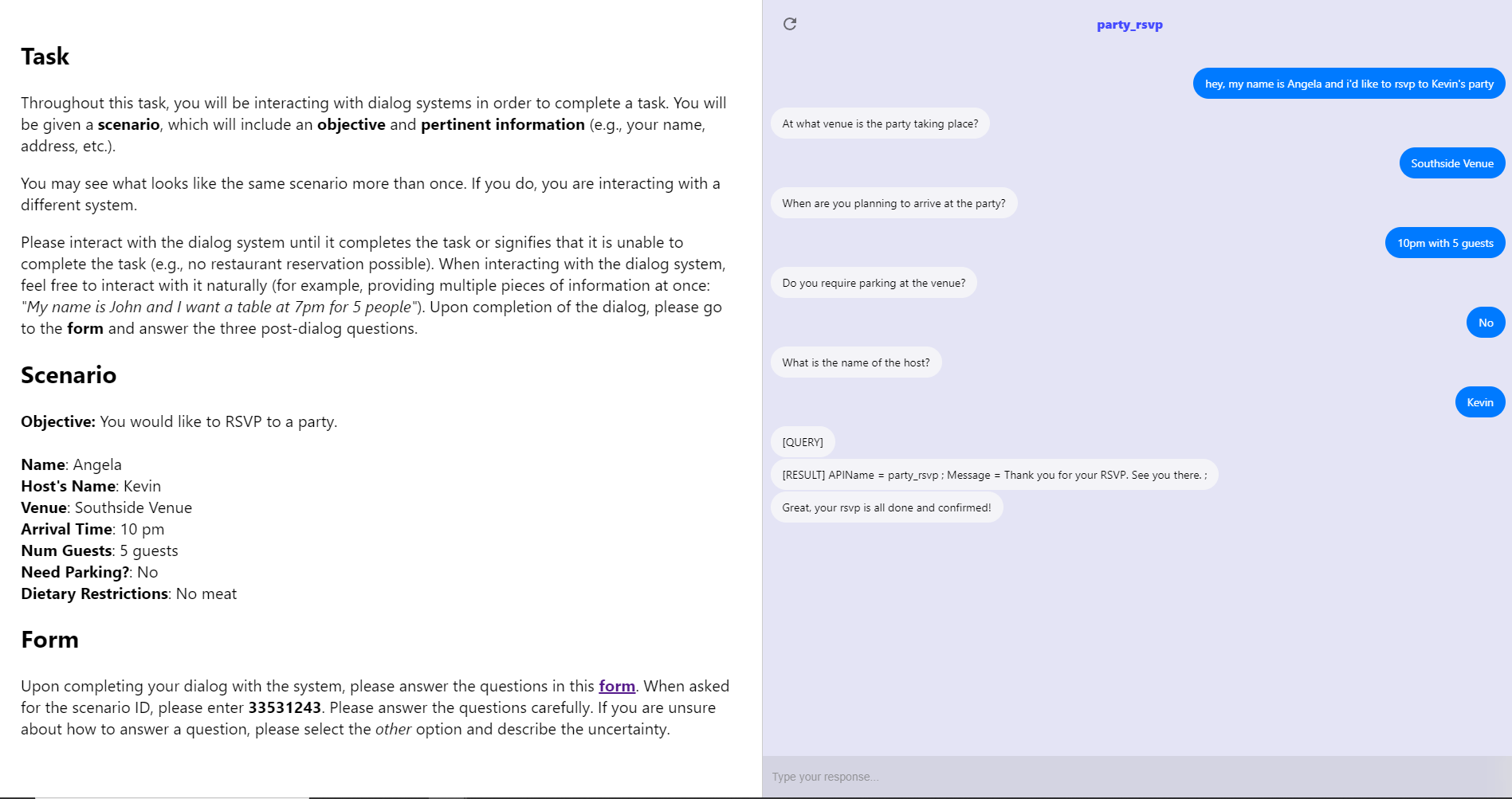}
    \caption{First example dialog that was used to evaluate AMT workers during the pre-screening. The correct answers for this example are: (1) the dialog is complete, (2) the system did not ask for all of the information (dietaray restrictions), (3) the system asked for redundant information (name of the host).}
    \label{fig:qual1}
\end{figure*}

\begin{figure*}
    \centering
    \includegraphics[width=\linewidth]{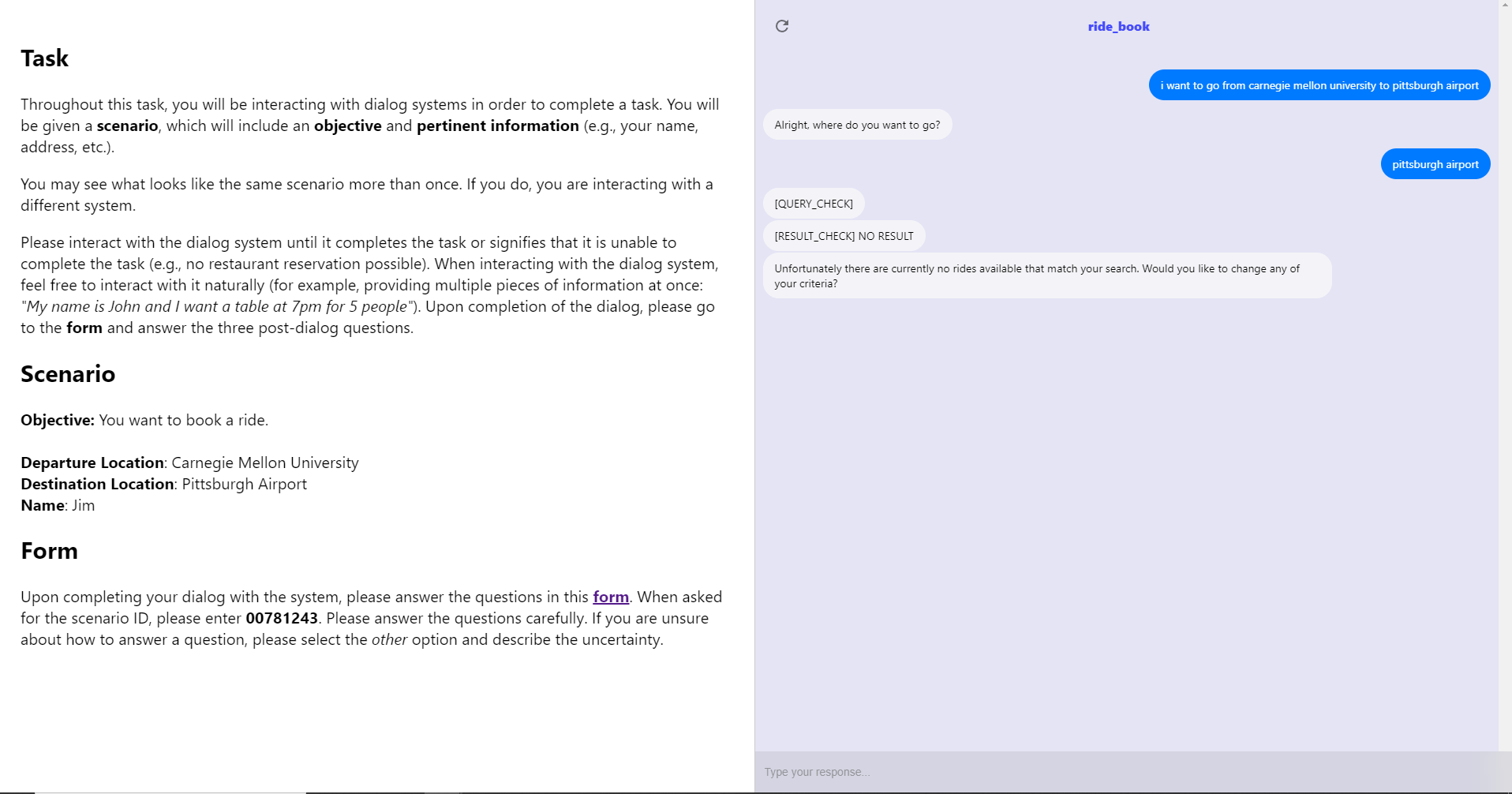}
    \caption{Second example dialog that was used to evaluate AMT workers during the pre-screening. The correct answers for this example are: (1) the dialog is complete, (2) the system did not ask for all of the information (name), (3) the system asked for redundant information (destination). The chosen slots in this image are sampled from the STAR corpus \citep{mosig2020star}, which is not necessarily the case for all of the scenarios.}
    \label{fig:qual2}
\end{figure*}

The interface that workers used to interact with a random dialog system are shown in Figures \ref{fig:qual1} and \ref{fig:qual2}. The post-dialog questions were answered in a Google Forms.

\begin{table*}
\centering
\renewcommand*{\arraystretch}{1.1}
{
\begin{tabularx}{\linewidth}{p{\linewidth}}
\toprule
This HIT is a pre-screening test for the Chatbot Evaluation HIT. Please read the instructions carefully and answer the questions below. \\

You will be qualified for the "Chatbot Evaluation" HIT, only if you correctly answer these pre-screening questions. You will be compensated \$0.75 for completing this pre-screening test. The Chatbot Evaluation task will pay \$3.25/HIT (\~10 minutes/HIT, ~600 total HITs), and will involve interacting with chatbots.\\
\\
For each dialog, you must answer three questions. Please read the instructions below carefully, before proceeding to the questions:\\

\midrule
\textbf{Question 1}: Did the system succesfully complete the dialog? This is asking whether the system successfully asked you all of the necessary information, understood your answers and performed a query ('[QUERY]') to find what you asked for. You should consider the dialog complete, even if the system indicated that they were unable to (for example) make you a reservation. A dialog might be incomplete, if (for example) the system fails to make a [QUERY] or repeatedly asks the same question.\\

\midrule
\textbf{Complete Dialog Example}: This dialog is considered complete because the system succesfully understood your instructions, but the restaurant is unavailable:\\

User: make me a reservation at Lucca's for 5 people at 7pm\\
System: \texttt{[QUERY]}\\
System: \texttt{[RESULT] [RestaurantName] is unavailable.}\\
System: [RestaurantName] is unable to make a reservation at that time, would you like to find another restaurant?\\

\midrule
\textbf{Incomplete Dialog Example}: This dialog is considered incomplete because the system fails to understand your instructions, even after multiple attempts:\\

User: make me a reservation at Lucca's for 5 people at 7pm\\
System: What is the name of the restaurant?\\
User: Lucca's\\
System: What is the name of the restaurant?\\
User: Lucca's\\
System: What is the name of the restaurant?\\
User: The name of the restaurant is Lucca's\\
System: What is the name of the restaurant?\\

\midrule
\textbf{Question 2}: Did the system ask you for all of the necessary information? If the system asked you for all of the information in the scenario, the answer to this question should be Yes. If the system neglected to ask you information even though it was provided in the scenario, the answer to this question should be No.\\

\midrule
\textbf{Positive Example}: In this dialog, the system received all of the information in the scenario before proceeding.\\

Scenario: Objective: Book a ride, Name: Sam, Departure Location: Union Grill, Destination Location: Central Park\\

User: i want a ride from Union Grill to Central Park\\
System: What is your name?\\
User: Sam\\
System: \texttt{[QUERY]}\\
System: \texttt{[RESULT] Departure=[DepartureLocation], Arrival=[ArrivalLocation], DriverName=[DriverName]}\\
System: Your ride is on its way.\\ \bottomrule
\end{tabularx}}%
\caption{Instructions for the pre-screening task. Continued in Table \ref{tab:qual_instructions2}}
\label{tab:qual_instructions}
\end{table*}

\begin{table*}
\centering
\renewcommand*{\arraystretch}{1.1}
{
\begin{tabularx}{\linewidth}{p{\linewidth}}
\toprule
\textbf{Negative Example}: In this dialog, the system failed to ask for the user's name before make a query.\\

Scenario: Objective: Book a ride, Name: Sam, Departure Location: Union Grill, Destination Location: Central Park\\

User: i want a ride from Union Grill to Central Park\\
System: \texttt{[QUERY]}\\
System: \texttt{[RESULT] Departure=[DepartureLocation], Arrival=[ArrivalLocation], DriverName=[DriverName]}\\
System: Your ride is on its way.\\

\midrule
\textbf{Question 3}: Did the system ask you for information that you had already provided it? If the system asked you for information that you had already provided, the answer to this question should be Yes. If the system did not ask for redundant information, the answer to this question should be No.\\

\midrule
\textbf{Positive Example}: In this dialog, the system did not ask for information that you had already provided.\\

User: i want a ride from Union Grill to Central Park\\
System: What is your name?\\
User: Sam\\
System: \texttt{[QUERY]}\\
System: \texttt{[RESULT] Departure=[DepartureLocation], Arrival=[ArrivalLocation], DriverName=[DriverName]}\\
System: Your ride is on its way.\\

\midrule
\textbf{Negative Example}: In this dialog, the system asked for the departure location, even though the user had already provided it.\\

User: i want a ride from Union Grill to Central Park\\
System: What is your name?\\
User: Sam\\
System: Where would you like to go?\\
User: Central Park\\
System: \texttt{[QUERY]}\\
System: \texttt{[RESULT] Departure=[DepartureLocation], Arrival=[ArrivalLocation], DriverName=[DriverName]}\\
System: Your ride is on its way.\\

\midrule
This HIT is a pre-screening test for the "Chatbot Evaluation" HIT. Please read the instructions carefully and answer the questions below. Succesfully passing this pre-screening will qualify you to do the Chatbot Evaluation HIT (\$3.25/HIT \& approx 10min/HIT \& approx 300 HITs).\\

Consider the scenario (left) and the dialog (right) in the image below. Answer the three questions according to the dialog.\\ \bottomrule
\end{tabularx}}%
\caption{Continuation of the pre-screening instructions.}
\label{tab:qual_instructions2}
\end{table*}

\subsubsection{Example Human-System Dialogs}

Example dialogs from the interactive human evaluation, with the \textsc{sam + lad} are shown in Table \ref{tab:ex_dial}.
\definecolor{dartmouthgreen}{rgb}{0.05, 0.5, 0.06}\begin{table*}
\definecolor{deepcarminepink}{rgb}{0.94, 0.19, 0.22}
\centering
\renewcommand*{\arraystretch}{1.1}
{
\begin{tabularx}{\linewidth}{p{\linewidth}}
\toprule
\textbf{Annotation:} {\color{dartmouthgreen} \textbf{Complete}, \textbf{Asked All}, \textbf{Avoided Redundancy}} \\
\midrule

\textbf{User:} my name is Morty Smith i dont have my account number but i need the balance, my dob is 9/19/2002 \\
\textbf{System:} What was your mother's maiden name? \\
\textbf{User:} sanchez \\
\textbf{System:} And what was the name of the pet you had as a child? \\
\textbf{User:} snuffles \\
\textbf{System:} And what was the name of the pet you had as a child? \\
\textbf{User:} snuffles \\
\textbf{System:} \texttt{[QUERY]} \\
\textbf{DB:} \texttt{[RESULT] APIName = bank\_balance ; BankBalance = [BankBalance]; BankName = [BankName] ; id = [NUMBER] ;}\\
\textbf{System:} Your current balance is [BankBalance] in credit. \\

\midrule
\textbf{Annotation:} {\color{dartmouthgreen} \textbf{Complete}}, {\color{deepcarminepink} \textbf{Did Not Ask All}}, {\color{dartmouthgreen} \textbf{Avoided Redundancy}} \\
\midrule

\textbf{User:} file a fraud report on account 68318931 \\
\textbf{System:} Right, and your PIN as well please. \\
\textbf{User:} 9503 \\
\textbf{System:} Could you provide a detailed fraud report now, please?
\textbf{User:} \$500 is missing \\
\textbf{System:} \texttt{[QUERY]} \\
\textbf{DB:} \texttt{[RESULT] APIName = bank\_fraud\_report ; Confirmation = Fraud report submitted successfully.} \\
\textbf{System:} Your report has been successfully submitted. We will have a look at the matter ASAP and will contact you with details in due course. \\

\midrule
\textbf{Annotation:} {\color{dartmouthgreen} \textbf{Complete}}, {\color{dartmouthgreen} \textbf{Asked All}}, {\color{deepcarminepink} \textbf{Redundant}} \\
\midrule

\textbf{User:} Hi, my name is Kevin and I yould like to plan a party on tuesday at 11 am on the southside venue with 85 guests \\
\textbf{System:} May I have your name, please? \\
\textbf{User:} Kevin \\
\textbf{System:} Would you like to order any specific kinds of food or drinks for your party? \\
\textbf{User:} Three course meal, with steak and a vegan option \\
\textbf{System:} \texttt{[QUERY\_CHECK]} \\
\textbf{DB:} \texttt{[RESULT\_CHECK] APIName = party\_plan ; Day = [Day]; Message = The venue is booked at that time. Try another meeting time or another venue. ; Time = [StartTimeHour] ; VenueName = [Name] ;} \\
\textbf{System}: I am sorry, but it looks like the [Name] is not available then. \\
\textbf{User:} ok\\
\bottomrule
\end{tabularx}}%
\caption{Interactive dialogs with \textsc{sam + lad}}
\label{tab:ex_dial}
\end{table*}

\subsection{Ablation}

In an effort to gain insight into the performance of \textsc{lad}, an ablation is carried out on next action prediction (Table \ref{tab:ablation}). To ensure a fair comparison, we generate 1000 dialogs with each approach. With reduced data, the performance of \textsc{SAM + lad} drops to \textbf{60.57}, which is still a strong improvement over prior work, without observing any human dialogs.

Removing the GPT-3 reformulation, i.e., using the seed data, results in an F-1 of 54.93. Removing the diversity of user behavior induced by Algorithm \ref{alg:cap} (and keeping the GPT-3 reformulation) results in a score of 48.91. This suggests that the diversity of user behavior is more important for facilitating effective learning of the necessary structural constraints (specifically the dialog policy) than the GPT-3 reformulation. 

The last row of Table \ref{tab:ablation} shows the results of leveraging an alternate prompt for GPT-3. Instead of generating \textbf{five} reformulations, this prompt only generates a single reformulation at a time. The reduced performance of 57.76 demonstrates the positive impact of our specific prompt. 

\begin{table}[!t]
\centering
\renewcommand*{\arraystretch}{1.1}
{
\begin{tabularx}{\linewidth}{l X}
\toprule
  \textbf{Model} & \multicolumn{1}{c}{\bf \textsc{F-1}} \\
  
  &  \\[-2ex] 
 \midrule
 
\textsc{SAM + \textsc{lad}}  &  \multicolumn{1}{c}{\textbf{60.57}} \\

~~~ -- \textit{reformulation} & \multicolumn{1}{c}{54.93} \\

~~~ -- \textit{seed diversity} & \multicolumn{1}{c}{48.91} \\

~~~ -- \textit{seed diversity}  -- \textit{reformulation} & \multicolumn{1}{c}{48.00} \\

~~~ -- \textit{five reformulation prompt} & \multicolumn{1}{c}{57.76} \\
\bottomrule
\end{tabularx}}%
\caption{Ablation experiments on the STAR corpus. To ensure a fair comparison, 1000 synthetic dialogs are generated with each approach. \textsc{SAM} was trained for 40 epochs on each synthetic dataset.}
\label{tab:ablation}
\end{table}

\end{document}